\documentclass{article}
\usepackage[T1]{fontenc}
\usepackage{ae}
\usepackage[utf8]{inputenc}
\usepackage{scrextend}
\usepackage{hyperref}
\usepackage{enumitem}
\usepackage{authblk}

\usepackage[margin=1.3in]{geometry}

\usepackage{amsmath}
\usepackage{amssymb}
\usepackage{amsthm}
\usepackage{amsfonts}
\usepackage{eucal}
\usepackage{algorithm}
\usepackage{algorithmic}
\usepackage{clrscode3e}

\usepackage{graphicx}
\usepackage{fancyhdr}
\usepackage{fancybox}
\usepackage{color}
\usepackage{fourier-orns}
\usepackage{wrapfig}
\usepackage{caption}
\usepackage{subcaption}
\usepackage{lmodern,textcomp}
\usepackage{hhline}
\usepackage{array}
\usepackage{makecell}

\usepackage{footnote}
\usepackage{tablefootnote}
\usepackage[symbol]{footmisc}

\usepackage[printwatermark]{xwatermark}
\usepackage{xcolor}
\usepackage{graphicx}
\usepackage{lipsum}

\usepackage{chngcntr}
\counterwithin{table}{section}
\counterwithin{figure}{section}

\usepackage{mathtools}
\DeclareMathAlphabet{\mathcal}{OMS}{cmsy}{m}{n}
\DeclarePairedDelimiter\ceil{\lceil}{\rceil}

\makeatletter

\newcommand*{\runinsubsection}{%
  \@startsection{subsection}%
  {2}
  {\z@}
  {-3.25ex\@plus -1ex \@minus -.2ex}
  {-1.5em \@plus -.1em}
  {\normalfont\large\bfseries}
}

\makeatother

\usepackage{csvsimple}

\DeclarePairedDelimiter\abs{\lvert}{\rvert}%
\DeclarePairedDelimiter\norm{\lVert}{\rVert}%
\makeatletter
\let\oldabs\abs
\def\abs{\@ifstar{\oldabs}{\oldabs*}}
\let\oldnorm\norm
\def\norm{\@ifstar{\oldnorm}{\oldnorm*}}
\makeatother

\numberwithin{equation}{section}

\title{Churn prediction in online gambling}
\author[a]{Merchie Florian\footnote{Corresponding author. Tel: +32494460103 \\ Email addresses: \href{florian.merchie@uliege.be}{florian.merchie@uliege.be} (F. Merchie), \href{fmerchie@uliege.be}{dernst@uliege.be} (D. Ernst)}}
\author[a,b]{Ernst Damien}
\affil[a]{Department of Electrical Engineering and Computer Science, ULiège}
\affil[b]{LTCI, Telecom Paris, Institut Polytechnique de Paris}
\date{}

\begin{document}

\maketitle

\begin{abstract}
    In business retention, churn prevention has always been a major concern. This work contributes to this domain by formalizing the problem of churn prediction in the context of online gambling as a binary classification task. We also propose an algorithmic answer to this problem based on recurrent neural network. This algorithm is tested with online gambling data that have the form of time series, which can be efficiently processed by recurrent neural networks. To evaluate the performances of the trained models, standard machine learning metrics were used, such as accuracy, precision and recall. For this problem in particular, the conducted  experiments allowed to assess that the choice of a specific architecture depends on the metric which is given the greatest importance. Architectures using nBRC favour precision, those using LSTM give better recall, while GRU-based architectures allow a higher accuracy and balance two other metrics. Moreover, further experiments showed that using only the more recent time-series histories to train the networks decreases the quality of the results. We also study the performances of models learned at a specific instant $t$, at other times $t^{\prime} > t$. The results show that the performances of the models learned at time $t$ remain good at the following instants $t^{\prime} > t$, suggesting that there is no need to refresh the models at a high rate. However, the performances of the models were subject to noticeable variance due to one-off events impacting the data. 
\end{abstract}

\section{Introduction}

In a business-to-consumer context, customer retention has always been considered as a key issue. Retaining a loyal customer base is crucial for the longevity of a company. However, enterprises also face a regular loss of clients, even those who have a loyal customer base. We usually call this loss 'churn', and its prevention is a major business challenge. The work proposed in \cite[Ascarza et al. 2016]{ascarza2016perils} explains how the behaviour of service providers can be modified by a rise of churn. This work also studies the effects of these changes and shows that usual offers such as attractive pricing plans can sometimes increase churn rather than decreasing it. The recent progress made in data science has eased the anticipation of churn events. By using customers' activity history, it is possible to identify different patterns associated to those who are the most likely to leave. This issue was addressed in \cite[Wei et al. 2002]{wei2002turning}. That work proposed a data mining technique able to detect such behaviors in the context of mobile services. However, in non-contractual business-to-consumer sectors, such as online gambling, the activity patterns are much more diverse. Indeed, the customers do not have, for instance, any monthly fee to pay and therefore will not feel obliged to use the service. In such contexts, churn detection becomes even more difficult to achieve. To solve such problems, more advanced technologies for analysing customer activity patterns, which can be represented as time series, can be used. \\

From a machine learning perspective, the analysis of time series for prediction has been an important research topic for a very long time. Among the many works in the field, several mention techniques using neural networks, such as \cite[LeCun et al, 1995]{lecun1995convolutional} and \cite[Zhang et al. 2003]{zhang2003time}. Those techniques have gained in popularity with the advent of recurrent neural networks (RNN), which have greatly increased the efficiency of developed models in this domain. Their ability to capture the temporal relationships within data is a major asset for prediction tasks requiring memory. Several research works, such as in \cite[Connot et al. 1994]{connor1994recurrent}, \cite[Graves et al. 2013]{graves2013generating}, and \cite[Che et al. 2018]{che2018recurrent}, proposed novel and robust solutions to time series-based prediction tasks using RNNs.
This places an emphasis on the relevance of using such techniques for solving business problems where customers' activity history can be described as time series.\\

In this work, we propose an accurate formalization of the churn prediction problem as a specific instance of a binary classification task. Then, we present an algorithmic solution to this problem applied to online gambling. This algorithm is based on the training of an RNN-based model whose goal is to predict churn occurrences over a given number of coming days. More particularly, at some time $t$, the model was learned from past histories. It was then used to predict the probability of churn of the different online players at this time $t$, over a given time horizon $T_{pred}$. We also assessed the ability of the model learned at time $t$ to predict the churn probability over the same horizon $T_{pred}$, but starting at other instants $t^{\prime} > t$. These data were collected from \textit{Gaming1}, a Belgian online gambling company. \\

This paper is organized in different sections. Section \ref{sec:related_work} presents an overview of the research work performed in churn prediction. Section \ref{sec:problem_statement} formalizes the statement of the churn prediction problem. Section \ref{sec:methodology} details the methodology of the proposed solution in terms of hypothesis space, loss function, and optimization algorithm. Section \ref{sec:experiments} details the different conducted experiments and their results. Section \ref{sec:conclusion} discusses the conclusion and future work perspectives around this paper. 

\section{Related work}
\label{sec:related_work}

Formalising a churn prediction problem is no small matter. This depends on both the churn definition and the evaluation metrics used for assessing the performances of the solution to the problem. The research paper \cite[Datta et al.]{datta2000automated} was the first to address a churn prediction problem with machine learning techniques. They introduced an automatized procedure applied to churn prediction in cellular phone services. Churn was defined as a per-customer monthly indicator based on the use of the services. Regarding other activity sectors, the research work of \cite[Buckinx et al., 2005]{buckinx2005customer} proposed a solution able to identify the potential defection of loyal customers in the insurance domain. Clients were segmented according to business criteria. Churn was then associated to loyal customers' risk of switching to a lower loyalty group. In the sector of digital music libraries, the research paper \cite[Lai et al., 2014]{lai2014analysis} formulated churn as a situation where a user does not download any music for a year. They aimed at identifying the periods in a user's life cycle where the churn risk was the highest and the customer groups associated to this risk. In the online gambling sector, the work proposed in \cite[Coussement et al., 2013]{coussement2013customer} evaluated the possibilities of predicting players' churn based on their activity history. They formalised churn as a four-month period without activity. In \cite[Suh et al, 2016]{suh2016customer}, churn was customized to each player by relying on how frequent and recent their activity patterns were. More specifically, a user was considered as a churner when the number of days since their last activity was greater than a value depending on the mean and standard deviation of the number of days between each activity. Concerning the other component of the problem statement, i.e., the performance metrics, the research works addressing churn prediction can be split into two categories. The first category, such as the research detailed in \cite[Buckinx et al., 2005]{buckinx2005customer} and \cite[Suh et al., 2016]{suh2016customer}, relies on the accuracy and confusion, i.e., standard machine learning metrics, to evaluate their solution. On the other hand, the second category, such as \cite[Coussement et al., 2013]{coussement2013customer} for instance, chose metrics specifically designed for churn detection. Among those, we can cite the top decile lift (TDL). To compute this value, churn probabilities need to be associated to customers. Then, customers are sorted according to this probability and the TDL consists of the ratio between the proportion of churners in the top 10\% of the ordered customers and the proportion of churners in the entire data set. Globally, we can observe the complexity of providing a unique definition for the problem of churn prediction, which would be applicable in any domain, which is mainly due to the diversity of business concerns in each activity sector. \\

Regarding the different machine learning techniques that were tested for churn prediction, the use of decision trees seems quite common, see for example \cite[Suh et al, 2016]{suh2016customer} and \cite[Coussement et al., 2013]{coussement2013customer}. In particular, we note that in \cite[Coussement et al., 2013]{coussement2013customer}, the performances of tree-based techniques were assessed on online sports betting data. The results showed that an ensemble of trees not only surpassed single-tree techniques, but also, for both, surpassed the equivalent of their Generalized Additive Model (GAM), a particular class of linear, equivalents. In the Chinese bank domain, the research paper of \cite{xie2009customer} showed that classical tree-based models were outperformed by a novel supervised learning method, called improved balance random forests. The model was also successfully challenged against other state-of-the-art methods such as support vector machine (SVM) and artificial neural networks (ANNs). ANNs were also used by \cite[Buckinx et al., 2005]{buckinx2005customer}, and shown to outperform random forests and logistic regression in the insurance sector. Research work \cite[Tsai et al., 2009]{tsai2009customer} introduced new methods using a combination of ANN and self-organizing maps (SOM) for churn prediction in American telecommunication services. The research paper of \cite[Farquad et al., 2014]{farquad2014churn} introduced a hybrid approach mixing SVM and rules generated using naive Bayes tree for customer retention in the bank credit card sector. The inability to scale up to large data sets was the main weakness of this method. Some unsupervised techniques were also evaluated, such as in \cite[Lai et al., 2014]{lai2014analysis}, where activity-based clustering was performed in order to prevent churn in online digital libraries. \\

\section{Problem statement}
\label{sec:problem_statement}

This section aims at defining the components of the problem statement, regarding churn indicator first, and loss function afterwards. In business terms, churn corresponds to a regular loss of customers, who stop any activity for a sufficiently long time period. This period can be chosen in an arbitrary way, depending on the activity sector. In online gambling, any customer can perform various activities, such as casino playing, sports betting, making deposits or withdrawals. These activities can be recorded and aggregated according to some arbitrary time step, and expressed as a multi-dimensional time series per customer. More formally, for any player $p \in \mathcal{P} = \{1, 2, \dots, |\mathcal{P}|\}$, these multi-dimensional time series can be formulated as $\mathrm{X}^p= \mathrm{x}_0^p, \mathrm{x}_1^p, \mathrm{x}_2^p \dots $, with $\mathrm{x}_i^p \in \mathbb{R}^n$, and where $n \in \mathbb{N}_0$ is the number of features used to describe the data. Among these features, only a subset of them is relevant to define the churn. This subset is denoted as $\mathrm{X}_c^p \subset \mathrm{X}^p$ and contains the same features for any player $p$. For each vector $\mathrm{x}^p \in \mathrm{X}_c^p$, we have $\mathrm{x}^p \in \mathbb{R}^{n_c}$, where $n_c < n$. For these activity features, we define a minimal number of days $T_c \in \mathbb{N}_0$ during which they must stay equal to 0 for churn to occur. Following this condition, for some player $p$, we can define the churn variable $c_t^p$ at some time $t$:
\begin{equation}
\label{eq:churn_def}
c_t^p = \text{max}(0, 1 - \ceil*{\sum_{i = t - T_c + 1}^t \sum_{j = 0}^n \abs{x_{i,j}^p}}),~\forall t > t-T_c, \mathrm{x} \in \mathrm{X}_c^p  
\end{equation}
where $c_t^p = 1$ (resp. $c_t=0$) stands for churn occurrence (resp. non churn). Then, we can define, for player $p$ at time $t$, the churn indicator $y_t^p$, which corresponds to a binary flag expressing whether player $p$ will churn at least once in the next $T_{pred}$ time steps after time $t$: 
\begin{equation}
    \label{eq:output_def}
    y_t^p = \text{min}(1, \sum_{i=t+1}^{t+T_{pred}}\!\!c_i^p),~\forall t \in \mathbb{N}_0.
\end{equation}
Having $y_t^p = 1$ (resp. $y_t^p = 0$) means player $p$ will churn (resp. will be retained) over the time horizon $T_{pred}$ starting at time $t$. In churn detection, we want to be able to predict this target value at time $t$ for any player $p$ having an activity history $i^p_t \in X^p_{0:t}$, where $X^p_{0:t}$ denotes the subset of the players' time series going from time $0$ to $t$. In this paper, we seek to learn a function $h_t$ that gives the probability of having a value of the churn indicator $y_t^p = 1$ for any player $p$, based on such a trajectory $i^p_t$. This corresponds to a binary classification problem, which can be solved using machine learning techniques. \\

The quality of such a function $h_t$ can be evaluated using a loss function defining the distance between the probability it outputs and the churn indicator $y_t^p$. In this paper, we choose the binary cross-entropy $H$ as loss function, which is most commonly used in classification, expressed as the following:
\begin{equation}
\label{eq:binary_cross_entropy}
H(y_t^p, h_t(i^p_t)) = -(y_t^p \log h_t(i^p_t) + (1-y_t^p)\log(1-h_t(i^p_t))).
\end{equation}
In practice, we learn this function $h_t$ at every time step $t$ and select it among a hypothesis space $\mathcal{H}$. The next section introduces and discusses the machine learning methodology chosen in this paper. \\

\section{Methodology}
\label{sec:methodology}

This section provides the formalization of the hypothesis space and the loss function used in the proposed algorithm. Then, it explains how the optimization procedure navigates through the hypothesis space to select its best function.  

\subsection{Hypothesis space}
\label{sec:hypothesis_space}

The learned function $h_t$ introduced in Section \ref{sec:problem_statement} is selected among some hypothesis space $\mathcal{H}$. In this work, every $\mathcal{H}$ that is used, contains a set of functions that output a probability based on an activity history $i^p_t$, for some player $p$. In particular, such a set consists here of all the neural networks corresponding to an architecture containing RNNs and feedforward networks, taking time series as inputs. An architecture is defined by its sequence of layers, where each layer depends on a set of parameters. A feedforward layer is fully specified by its number of neurons, weights, and activation function. The same elements are also used for defining recurrent layers, that, contrary to feedforward networks, also have a hidden state. This hidden state $\mathrm{h}_t = f(\mathrm{h}_{t-1}, \mathrm{x}_t;\theta)$, where $\mathrm{h}_0$ is a constant and $\theta$ are the parameters of the network, can be used by the network to store past information of the time series. A syntax enabling one to easily specify such complex ANN architectures  was proposed by \cite[Vecoven et al., 2020, Appendix 3]{vecoven2020introducing}. With this syntax, a layer is represented by an acronym referring to the type of neurons it is made of. This acronym is followed by the symbol $\circlearrowright$ when it is a recurrent layer. The acronym also takes as subscript the number of neurons in a layer. A $\rightarrow$ symbol is used to highlight the connections between the different layers, e.g.:
\begin{equation}
    \label{eq:synthax_example}
     LSTM_{10} \!\!\, \circlearrowright \: \rightarrow LSTM_{10} \!\!\, \circlearrowright \: \rightarrow sigm_1,
\end{equation}
defines a three-layer neural network. The first two are recurrent layers having 10 Long-Short Term Memory (LSTM,  \cite[Hochreiter et al., 1997]{hochreiter1997long}) cells each. The last one is a single neuron using a sigmoid activation function. Using this syntax, we can provide the generic architecture that implicitly defines the hypothesis space $\mathcal{H}$ that will be used in our experiments:
\begin{equation}
    \label{eq:hypothesis_space_1}
     Rec_{N} \!\!\, \circlearrowright \: \rightarrow Rec_{M} \!\!\, \circlearrowright \: \rightarrow sigm_1,
\end{equation}
where $Rec_{N} \!\!\, \circlearrowright$ denotes a $N$-neuron recurrent layer and $Rec$ corresponds to either a Gated Recurrent Unit (GRU, \cite[Cho et al., 2014]{cho2014learning}), LSTM, or neuromodulated Bistable Recurrent Cell (nBRC, \cite[Vecoven et al., 2020]{vecoven2020bio}).

\subsection{Optimization procedure}
\label{sec:opti}

To find the best function $h_t$ at time $t$ among the hypothesis space, the machine learning algorithm aims at minimizing an empirical loss function $L_t$ over a learning data set $LS_t$. This learning data set is composed of pairs of the type $(i_{t^{\prime}}^p,y_{t^{\prime}}^p)$ where $i_{t^{\prime}}^p$ is the history trajectory of an eligible player $p$ available at time $t^{\prime}$ and $y_{t^{\prime}}^p$ is the churn indicator for player $p$ at time $t^{\prime}$. The loss function is based on Equation \ref{eq:binary_cross_entropy} and is equal to the averaged sum of the binary entropy losses computed over the training set, namely: 
\begin{equation}
\label{eq:loss_function_dataset}
L_t(LS_t) = \frac{1}{|LS_t|}\sum_{p = 1}^{P} \sum_{t^{\prime}=\max(T_0, t_0^p)}^{t-T_{pred}}H(y_{t^{\prime}}^p, h_t(i^p_{t^{\prime}})).
\end{equation}
A few remarks ought to be made concerning Equation \ref{eq:loss_function_dataset}. First, the upper bound set for $t^{\prime}$ comes from Equation \ref{eq:output_def}, that defines the churn indicator. This equation stresses that, knowing only $i_t$ when learning $h_t$, churn indicators can be computed for $t^{\prime} <= t - T_{pred}$ at most. Also, we faced a trade-off when choosing the lower bound of $t^{\prime}$. If we had set it to $t - T_{pred}$, we would have used only the most recent information, but we would probably have missed some relevant data contained in older pairs. The other way around, nothing prevents us from fixing this lower bound to 0. The reason for not doing this is twofold: (i) it may happen that, at the beginning of the trajectories, very little information is available to carry out a good prediction, which, by selecting data pairs for very small values of $t^{\prime}$, would only correspond to adding noise to the loss function (ii) we prefer having in the training set pairs which correspond to churn indicators that are not too old. Indeed, even though Equation \ref{eq:loss_function_dataset} implicitly assumes that churn patterns are somehow stationary with time, nothing prevents customers from modifying their behaviour regarding churn. For example, we could imagine the release of a much more addictive game. The customers playing that game would then be very likely to have their churn patterns modified. Finally, this leads us to define two parameters: a constant $T_0$, which is not player dependent, and a constant per player $t_0^p$, whose maximum corresponds to the lowest time step at which player $p$'s data are considered for building $LS$. According to these remarks, player $p$ intervenes through $t - T_{pred} - \max(T_0, t_0^p) + 1$ terms in the loss function. The value of $T_0$ will depend on the conducted experiments. Additionally, in our simulations, we will restrict the size of the time series to its last $T_h$ elements. \\

The optimization procedure enabling the selection the hypothesis function $h_t$ at time $t$ is formulated as follows:
\begin{equation}
    \label{eq:minimization_procedure}
    h_t \in \underset{h_t \in \mathcal{H}}{\text{argmin}}~ L_t(LS_t).
\end{equation}
Finding $h_t$ is thus equivalent to choosing the point of the hypothesis space minimizing $L_t$ for the training data available at time $t$. In the scope of ANN, the problem defined by Equation \ref{eq:minimization_procedure} is most commonly solved using gradient descent algorithms. Such algorithms consist of approximating the gradient of a given function and iteratively following the opposite direction of the gradient in order to find a local minimum of that function. Over the past years, several variants of this algorithm were designed in order to fit different ANN architectures. Among these, we chose RMSprop for this work. It was introduced in two steps by \cite[Hinton et al, 2012]{hinton2012neural}. They first discussed the variability of the magnitude of the gradient in function of the weights of the DNN and of the progression of the learning to introduce the Rprop algorithm. Then, it was adapted in order to be compatible with mini-batch learning. As any gradient descent algorithm, RMSprop can be tuned according to several parameters: mini-batch size $size_{batch}$, number of epochs $n_{epochs}$ and learning rate $\alpha$. It also has a specific parameter $\rho$, the decay factor for the gradient. Each parameter is specified for every experiment, and some of them will be discussed in Section \ref{sec:experiments}. Finally, we note that, for solving the optimization problem \ref{eq:minimization_procedure} in our experiments, we unroll the time-series over all its selected elements for computing the gradient. \\

\section{Experiments}
\label{sec:experiments}

In this section, we test our proposed approach on data from an online gaming company, \textit{Gaming1}. In this company, each transactional activity is recorded into a database, along with the exact time at which it occurred. For this work, we choose a time step of 1 day. As a result, each time-series element consists of a multidimensional vector aggregating all considered features over a whole day. These features, which make up the sets $X^p$ and $X^p_c$, are listed in Table \ref{tab:features} \footnote{The Gross Gaming Revenue (GGR) denotes the revenue earned by the company from the player's activity. In practice, it is computed by subtracting the winnings from the stakes for each player.}. The number of website connections correspond to the number of times the player logged into the website of the company, independently of their subsequent activity. \\

\begin{table}[h]
    \centering
    \begin{tabular}{|c|c|c|}
        \hline
        \textbf{Type} &  \textbf{Feature} &  \textbf{Belongs to the set(s)} \\ \hline
        Casino & Number of casino plays & $X^p$ and $X^p_c$ \\ \hline
        Casino & Casino Gross Gaming Revenue (GGR)  & $X^p$ \\ \hline
        Sport betting & Number of sport betting tickets & $X^p$ and $X^p_c$ \\ \hline
        Sport betting & Sport betting GGR  & $X^p$  \\ \hline
        Transactions & Number of deposits & $X^p$ and $X^p_c$ \\ \hline
        Transactions & Number of withdrawals & $X^p$ \\ \hline
        Logging & Number of website connections & $X^p$ \\ \hline
        Retention & Number of days since last active day & $X^p$ \\ \hline
    \end{tabular}
    \caption{Features aggregated from transactional logs}
    \label{tab:features}
\end{table}

In this company, the retention services identify a customer as a churner when the number of days since his last active day is least equal to $T_c = 35$. A day is considered as active when the player performs at least one gaming session, sport betting ticket, or deposit within this day. Concerning churn prediction, \textit{Gaming1} wants to assess their customers' churn probability within the next $T_{pred} = 30$ days. This duration has been chosen to allow retention campaigns to be scheduled sufficiently in advance. Moreover, we set the minimum time step per player $t_0^p$ to the $60^{\text{th}}$ day after player $p$'s registration date for building the training set. Before this day, the rules used by \textit{Gaming1} for retention are different. Therefore, we have also chosen to use the last $T_h = 60$ elements of player $p$'s time series when collecting data pairs $(i_{t^{\prime}}^p,y_{t^{\prime}}^p)$. Table \ref{tab:static_params} gathers the values of these parameters, which remain the same for all conducted experiments. The collected data comprise the activity history of $10~000$ gambling players from the $1^{\text{st}}$ of January 2017 to the $1^{\text{st}}$ of June 2021. For players registered after $1^{\text{st}}$ of January 2017, the missing parts of the time-series are filled with 0. \\ 

\begin{figure}[h]
    \centering
    \includegraphics[width=0.9\textwidth]{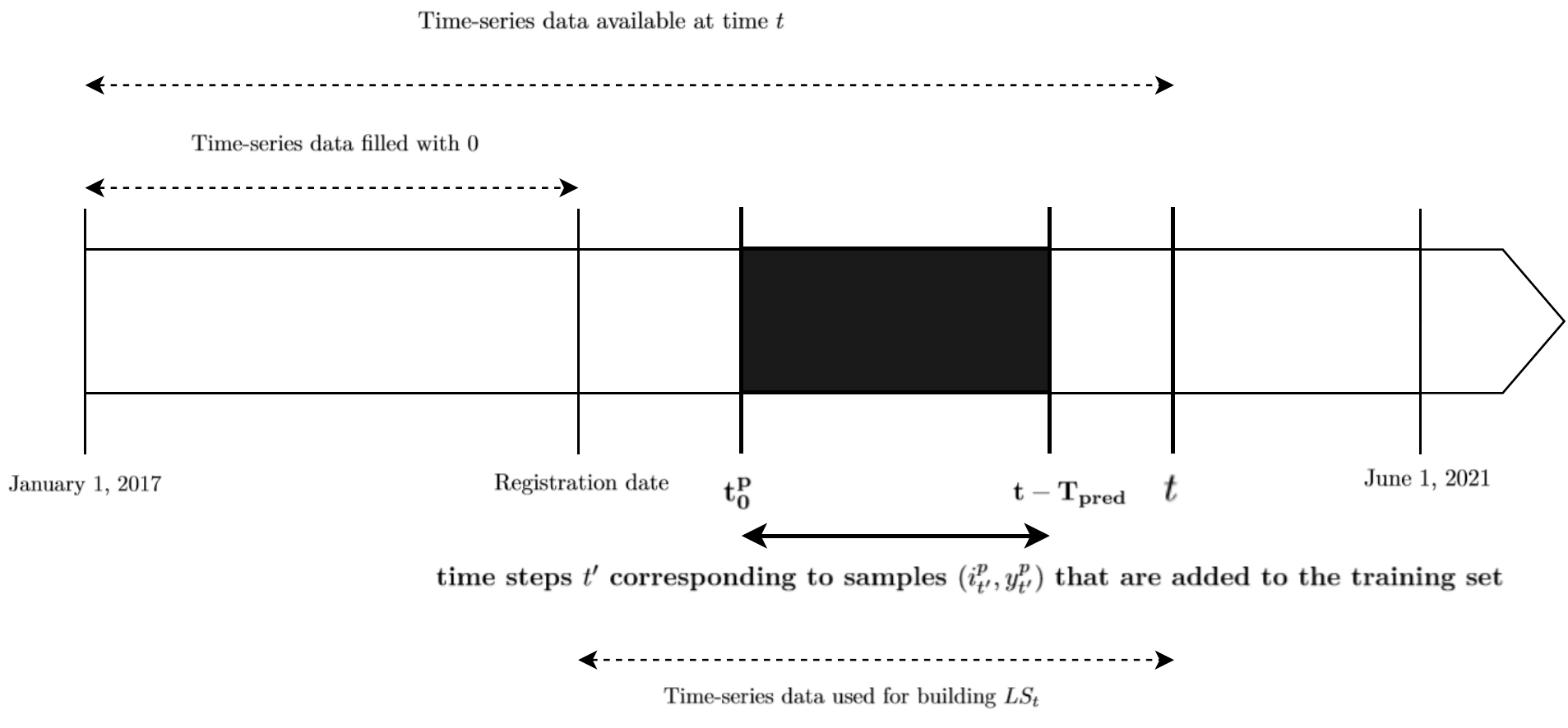}
    \caption{Timeline displaying the time steps $t^{\prime}$ to which correspond pairs $(i_{t^{\prime}}^p,y_{t^{\prime}}^p)$ used for building the training set at time $t$.}
    \label{fig:data_training}
\end{figure}

We now explain how we build our training set $LT_t$ at time $t$ from the collected players' histories. This will be composed of all the pairs $(i_{t^{\prime}}^p,y_{t^{\prime}}^p)$ that can defined at time $t$ and that also satisfy the following conditions, that we call activity conditions: 
\begin{enumerate}[label=(\roman*)]
    \item $t^{\prime} \ge \max(T_0, t_0^p)$. 
    \item player $p$ must have at least 1 active day within the 30 days that precede $t^{\prime}$.
\end{enumerate}
We note that, from these conditions, we can derive an upper bound on the number of samples related to some player $p$ used in $LS_t$, which is equal to $t - T_{pred} - \max(T_0, t_0^p) + 1$. As way of summary, Figure \ref{fig:data_training} displays a timeline explaining which data pairs are used to build $LS_t$ for some player $p$. More specifically, the black-coloured block corresponds to the time steps $t^{\prime}$ that feed the learning set $LS_t$ with pairs $(i_{t^{\prime}}^p,y_{t^{\prime}}^p)$. 

\subsection{Standard experiment}
\label{sec:standard_experiment}

In this section, we use the learning set $LS_t$ previously characterized for computing $h_t$ with $t$ corresponding to the $1^{\text{st}}$ of January 2020. This model $h_t$ will be used for forecasting whether some player $p$ will churn in the 30 days following $t$, or equivalently to predict the value of $y^p_{2020/01/01}$. We note that players for which this evaluation will take place at time $t$, are those who have registered for at least 60 days and have at least one active day in the 30 days preceding $t$, two conditions that are similar to the activity conditions introduced before. We obtain a test set $TS_{2020/01/01}$ containing $1~924$ samples, among which $\sim 22$\% have a value of $y_{t}^p$ equal to 1. Those correspond to a churn event. The rest, having a value $y_{t}^p$ of 0, is associated with a retained event. Concerning the training data, $LT_{2020/01/01}$ is composed of $793~779$ elements. Among these pairs, $\sim 22$\% (resp. $\sim 78$\%) are churn (resp. retained) samples. We use the precision and recall of the churn class, along with the accuracy, to evaluate the performances of the model. These metrics are defined as: 
\begin{equation}
    \text{precision} = \frac{\text{TP}}{\text{TP}+\text{FP}},
\end{equation}
\begin{equation}
    \text{recall} = \frac{\text{TP}}{\text{TP}+\text{FN}},
\end{equation}
\begin{equation}
    \text{accuracy} = \frac{\text{TP}+\text{TN}} {\text{TP}+\text{FP}+\text{TN}+\text{FN}},
\end{equation}
where TP is the number of true positives, FP the number of false positives, TN the number of true negatives, and FN the number of false negatives. Precision corresponds to the rate of correct predictions for the number of customers predicted as churners, while the recall denotes the rate of churn customers that are identified by the model. Results have been generated based on three random seeds and can be found in Table \ref{tab:standard_experiment_results}. We note that the accuracy is almost identical for the three types of RNN. The choice of the architecture depends instead on the objective we want to achieve in retention. When we need cost-effective campaigns, the architecture using nBRC layers is the best choice, as it obtains the best precision value for the churn event class. In contrast, LSTM layers allow for the detection of the highest number of churn customers by getting the best recall value for the churn class. When all metrics are equally important, GRU layers prevail. Indeed, it gets the best accuracy score and the second-best value for the other metrics. For further experiments, we decide to use the LSTM layers. 

\begin{table}[!t]
    \centering
    \begin{tabular}{|c|c|c|c|}
        \hline
        \textbf{Layer} & \textbf{Precision - Churn class} & \textbf{Recall - churn class} & \textbf{Accuracy} \\ \hline
        GRU & 0.636212 ± 0.001635 & 0.574519 ± 0.037549 & \textbf{0.836972 ± 0.003461} \\ \hline
        LSTM & 0.616472 ± 0.008382 & \textbf{0.611378 ± 0.061255} & 0.833853 ± 0.007375 \\ \hline
        nBRC & \textbf{0.644099 ± 0.006254} & 0.485577 ± 0.025440 & 0.830735 ± 0.002456 \\ \hline
    \end{tabular}
    \caption{Performances of the $h_{2020/01/01}$, for different RNN layers and random seeds. The complete list of parameters is available in Table \ref{tab:params_standard}.}
    \label{tab:standard_experiment_results}
\end{table}

\begin{figure}[!t]
    \centering
    \includegraphics[width=0.7\textwidth]{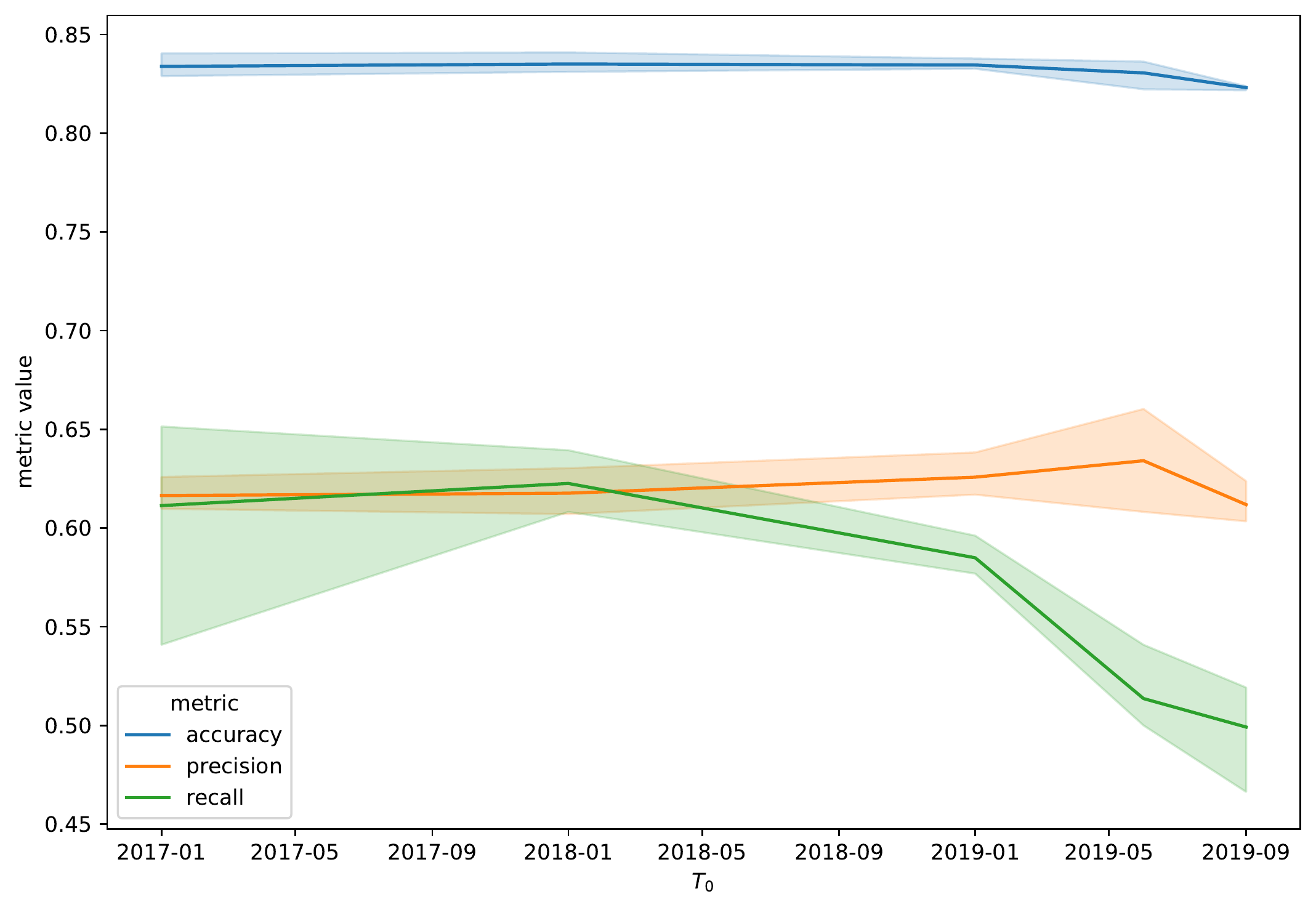}
    \caption{Performance metrics of $h_{2020/01/01}$, using LSTM layer. The metrics are computed for different values of $T_0$, leading to different training sets, and different random seeds. The complete list of parameters is available in Table \ref{tab:params_influence_t0}}
    \label{fig:influence_t0}
\end{figure}

\subsection{Influence of $T_0$}
\label{sec:influence_t0}

In this section, we want to study the impact of a variation of the value of the $T_0$ parameter on the performances of the model trained in the previous experiment. This parameter was introduced in Section \ref{sec:opti} and influences the number of samples that we can use per player to build the training data set. Lower values of $T_0$ lead to a data set with many samples, with the risk that some of them may be related to past situations not relevant for churn prediction at time $t$. In contrast, later values result in smaller data sets but with more-recent samples. With this experiment, we want to determine whether it is better to use a greater amount of data, or more-recent ones, to solve the problem stated in Section \ref{sec:problem_statement}. The different values of the parameters used for this experiment can be found in Table \ref{tab:params_influence_t0}. The results of the experiment are displayed in Figure \ref{fig:influence_t0} for the different values of $T_0$. Globally, we observe that $T_0$ has a relatively small impact on the precision and accuracy curves. However, we observe a progressive drop of the churn recall for values of $T_0$ greater than January 2018. This means using only recent data for building the training set decreases the number of churners detected by the model. This experiment suggests we are in a context where using only the most recent data is not favourable and that relatively low values of $T_0$ can be used. 

\subsection{Model robustness over time}
\label{sec:time_robustness}

\begin{figure}[!t]
    \centering
    \includegraphics[width=0.7\textwidth]{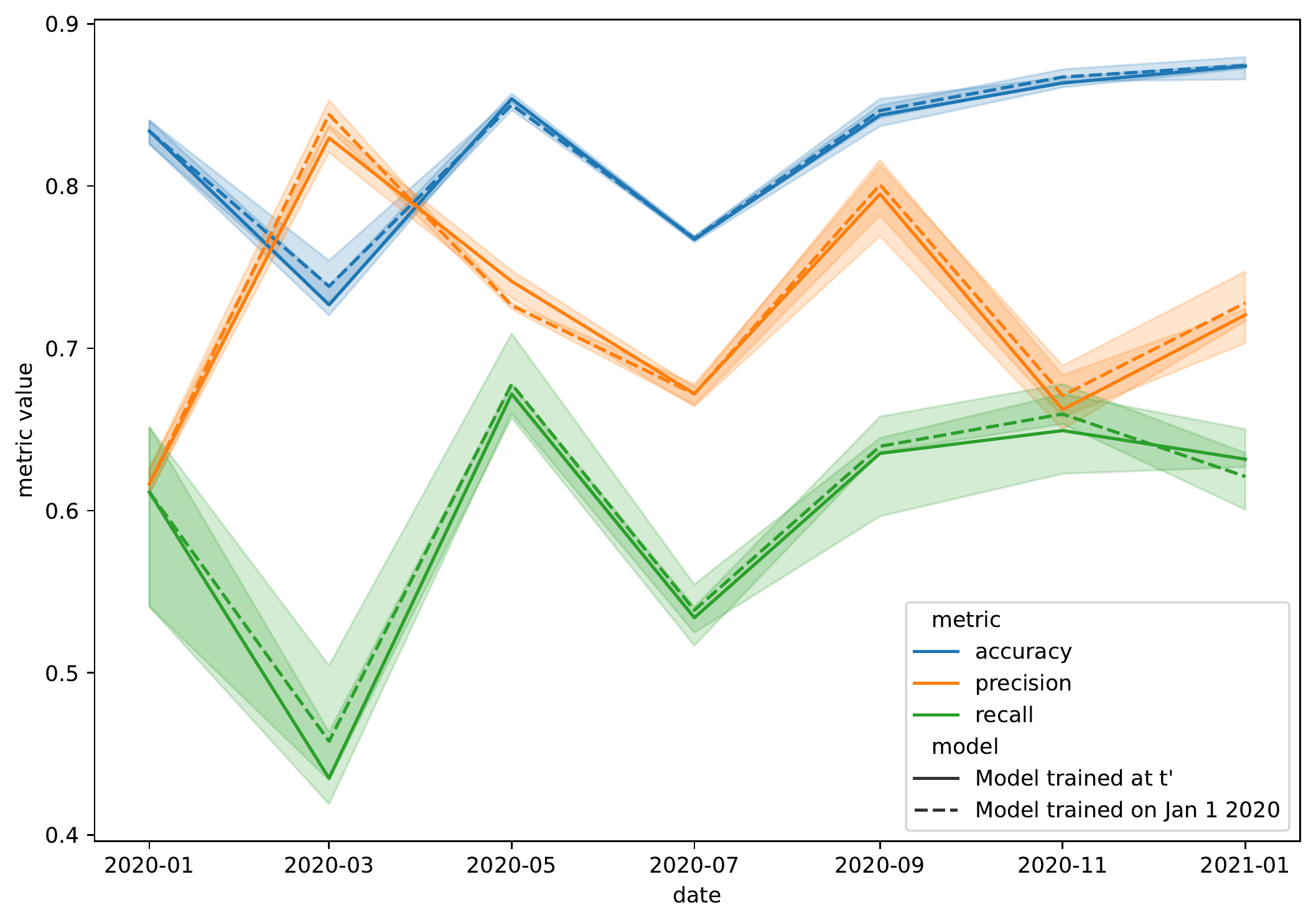}
    \caption{Performance metrics of the model learned on the $1^{\text{st}}$ of January 2020, using LSTM layer for several $LS_{t^{\prime}}$ after this date. The metrics are computed for different random seeds and compared to those obtained by models trained using $LS_{t^{\prime}}$. The complete list of parameters is available in Table \ref{tab:params_time_robustness}}
    \label{fig:time_time_robustness}
\end{figure}

In this last experiment, we want to evaluate the robustness over time of the model trained in the experiment detailed in Section \ref{sec:standard_experiment}. To do so, we compute the evolution of the performances of the model trained using $LS_{2020/01/01}$, i.e. $h_{2020/01/01}$, on several testing sets $TS_{t^{\prime}}$, with $t^{\prime}$ greater or equal to the $1^{\text{st}}$ of January 2020. Then, for each $t^{\prime}$, we train another model $h_{t^{\prime}}$ using the set $LT_{t^{\prime}}$, that is built using all the data pairs satisfying at $t^{\prime}$ the activity conditions mentioned in the introduction paragraph of Section \ref{sec:experiments}. Using the same metrics as in previous experiments, we evaluate the performances of $h_{t^{\prime}}$ at $TS_{t^{\prime}}$, for each chosen $t^{\prime}$. We select the different values of $t^{\prime}$ to uniformly cover the period going from the $1^{\text{st}}$ of January 2020, to the $1^{\text{st}}$ of January 2021, with a time step of two months between each $t^{\prime}$. The parameters used in the experiment are reported in Table \ref{tab:params_time_robustness}. The random seeds are the same as the previous two experiments. Figure \ref{fig:time_time_robustness} display the values of the different metrics for the different test dates. While both models offer similar performances, we may observe significant variations over time, for example, in the way the recall metric evolves. We conjecture that those could be related to specific major events experienced by the company. For instance, on the $1^{\text{st}}$ of March 2020, the company stopped offering several types of bonus for legal reasons. This may explain the strong discontinuity observed at that time. In particular, we note that the sharp rise in precision observed at that time might be a side effect of the recall drop. Indeed, such trends are probably due to the model making many fewer churn guesses than those that indeed occur at other time steps. As a consequence, only the most certain churn events are effectively predicted as they are by the model, causing these variations in recall and precision. Similarly, at the end of June 2020, a technical issue made the casino unreachable for several days. In both cases, the company faced events that most likely impacted their players' activity patterns. As a consequence, the past histories used to train the model were not relevant to predict churn events for these particular months. 

\section{Conclusion and perspectives}
\label{sec:conclusion}

As a conclusion, this study shows how to formalise the churn prediction problem and how recurrent neural networks can be used to solve it. Given a sufficient volume of data, we demonstrate that RNNs can achieve solid performances. Indeed, even though the occurrence of churn is relatively rare in the data used in the different experiments, the models still get honourable results at detecting this event. Moreover, by comparing the performances obtained in Section \ref{sec:standard_experiment}, we observe that the choice of the architecture could change according to the desired objectives of the retention campaigns. If we want to conduct cost-effective retention campaigns, nBRC is probably the best option, as it had the highest precision values for the churn class. When the top priority is to detect as many churn customers as possible, we would probably go for LSTM architectures as these the best recall for the churn events. Then, Section \ref{sec:influence_t0} shows that using only more-recent data to train the model can decrease the performances. In particular, the recall of the churn events starts to drop when the size of the time window used for training become lower than one year. Furthermore, with the experiment conducted in Section \ref{sec:time_robustness}, we concluded that, in our given context, a model could sustain competitive performances over time compared to other models trained at posterior time steps. However, we also highlighted some non-negligible sensitivity to one-off events that can impact the players' activity patterns for a short period of time. Eventually, there are still possibilities for future research work in this domain. The room for improvement in the detection of churn events is significant because of the rarity of such events. Techniques based on anomaly detection would be an interesting option to tackle this problem from another point of view. As an example, the research paper \cite{sundarkumar2015one} challenged one class SVM to credit card customers churn data set, showing it is possible to obtain decent results using anomaly detection methods in this domain. Also, a gambling company does not necessarily have the same across-the-board reason for retaining all its churning customers, as some may be more profitable than others. Combining the player value with the churn probability could help to increase the return on investment of retention campaigns.            

\runinsubsection*{Acknowledgements.} The authors would like to thank \textit{Gaming1} for providing the data used in the experiments of this paper, and \textit{Gaming1} data science team for their support and counsel for this work.  

\bibliographystyle{unsrt}
\bibliography{bibli}

\newpage

\appendix

\section{Parameter values}

\begin{table}[h]
    \centering
    \begin{tabular}{|c|c|c|}
        \hline
        \textbf{Parameter} & Description & \textbf{Value}  \\ \hline
        $T_c$ & Number of inactive days for churn & 35 \\ \hline
        $T_{pred}$ & Churn prediction period & 30 \\ \hline
        $T_h$ & Size of input time series in days & 60 \\ \hline
        $t_0^p$ & \makecell{Minimum time step for some player $p$'s \\ data pair to belong to $LS$}  & \makecell{60 days after \\ player $p$'s registration date} \\ \hline
    \end{tabular}
    \caption{Global parameters for all experiments}
    \label{tab:static_params}
\end{table}

\begin{table}[h]
    \centering
    \begin{tabular}{|c|c|c|}
        \hline
        \textbf{Parameter} & \textbf{Description} & \textbf{Value}  \\ \hline
        $N$ &  Number of neurons of the first recurrent layer  & 128 \\ \hline
        $M$ &  Number of neurons of the second recurrent layer  & 64 \\ \hline
        $T_0$ & Start of the time interval used to build $LS$  & 2017-01-01 \\ \hline
        date & Date of the experiment & 2020-01-01 \\ \hline
        $n_{epochs}$ & Number of epochs per training procedure  & 20 \\ \hline
        $size_{batch}$ & Size of sample batch for each training step  & 256 \\ \hline
        $\alpha$ & Learning rate of the optimizer used in the learning procedure & $10^{-3}$ \\ \hline
        $\gamma$ & Decay factor of the gradient & 0.9 \\ \hline
    \end{tabular}
    \caption{Parameter values for the standard experiment \ref{sec:standard_experiment}}
    \label{tab:params_standard}
\end{table}

\begin{table}[h]
    \centering
    \begin{tabular}{|c|c|c|}
        \hline
        \textbf{Parameter} & \textbf{Description} & \textbf{Value}  \\ \hline
        $N$ &  Number of neurons of the first recurrent layer  & 128 \\ \hline
        $M$ &  Number of neurons of the second recurrent layer  & 64 \\ \hline
        RNN &  Type of recurrent layer  & LSTM \\ \hline
        date & Date of the experiment & 2020-01-01 \\ \hline
        $n_{epochs}$ & Number of epochs per training procedure & 20 \\ \hline
        $size_{batch}$ & Size of sample batch for each training step  & 256 \\ \hline
        $\alpha$ & Learning rate of the optimizer used in the learning procedure & $10^{-3}$ \\ \hline
        $\gamma$ & Decay factor of the gradient & 0.9 \\ \hline
    \end{tabular}
    \caption{Parameter values for the experiment \ref{sec:influence_t0} studying the influence of $T_0$}
    \label{tab:params_influence_t0}
\end{table}

\begin{table}[h]
    \centering
    \begin{tabular}{|c|c|c|}
        \hline
        \textbf{Parameter} & \textbf{Description} & \textbf{Value}  \\ \hline
        $N$ &  Number of neurons of the first recurrent layer  & 128 \\ \hline
        $M$ &  Number of neurons of the second recurrent layer  & 64 \\ \hline
        RNN &  Type of recurrent layer  & LSTM \\ \hline
        $T_0$ & Start of the time interval used to build $LS$  & 2017-01-01 \\ \hline
        $n_{epochs}$ & Number of epochs per training procedure & 20 \\ \hline
        $size_{batch}$ & Size of sample batch for each training step  & 256 \\ \hline
        $\alpha$ & Learning rate of the optimizer used in the learning procedure & $10^{-3}$ \\ \hline
        $\gamma$ & Decay factor of the gradient & 0.9 \\ \hline
    \end{tabular}
    \caption{Parameter values for the time robustness experiment \ref{sec:time_robustness}}
    \label{tab:params_time_robustness}
\end{table}

\end{document}